\documentclass[journal]{IEEEtran}
\IEEEoverridecommandlockouts

\usepackage{booktabs}
\usepackage{graphicx}
\usepackage{caption}
\usepackage{float}
\usepackage{color}
\usepackage{multirow}
\usepackage{makecell}

\usepackage{cite}

\def\BibTeX{{\rm B\kern-.05em{\sc i\kern-.025em b}\kern-.08em
    T\kern-.1667em\lower.7ex\hbox{E}\kern-.125emX}}

\usepackage{array}
\newcolumntype{L}[1]{>{\raggedright\let\newline\\\arraybackslash\hspace{0pt}}m{#1}}
\newcolumntype{C}[1]{>{\centering\let\newline\\\arraybackslash\hspace{0pt}}m{#1}}
\newcolumntype{R}[1]{>{\raggedleft\let\newline\\\arraybackslash\hspace{0pt}}m{#1}}

\begin{document}

\title{Evaluating Fundus-Specific Foundation Models for Diabetic Macular Edema Detection}

\author{
    \IEEEauthorblockN{Franco Javier Arellano}
    , \IEEEauthorblockN{José Ignacio Orlando}
    \\
    \IEEEauthorblockA{\textit{Yatiris Group, PLADEMA Institute, UNICEN, Tandil, Argentina}}\\
    \IEEEauthorblockA{\textit{CONICET, Tandil, Argentina}}\\
    \IEEEauthorblockA{francoare@pladema.exa.unicen.edu.ar, jiorlando@pladema.exa.unicen.edu.ar}
}

\maketitle

\begin{abstract}
Diabetic Macular Edema (DME) is a leading cause of vision loss among patients with Diabetic Retinopathy (DR). 
While deep learning has shown promising results for automatically detecting this condition from fundus images, its application remains challenging due the limited availability of annotated data. 
Foundation Models (FM) have emerged as an alternative solution. 
However, it is unclear if they can cope with DME detection in particular.
In this paper, we systematically compare different FM and standard transfer learning approaches for this task. 
Specifically, we compare the two most popular FM for retinal images--RETFound and FLAIR--and an EfficientNet-B0 backbone, across different training regimes and evaluation settings in IDRiD, MESSIDOR-2 and OCT-and-Eye-Fundus-Images (OEFI). Results show that despite their scale, FM do not consistently outperform fine-tuned CNNs in this task. \textcolor{black}{
 In particular, EfficientNet-B0 consistently achieves competitive or superior performance across evaluation settings, with FLAIR being the most competitive foundation model, consistently outperforming RETFound. These findings suggest that FMs do not necessarily provide an advantage for fine-grained ophthalmic tasks such as DME detection, even after fine-tuning, highlighting lightweight CNNs as strong baselines in data-scarce environments.}

\end{abstract}

\section{Introduction}

\begin{figure}[t!]
  \centering
  \includegraphics[width=0.99\columnwidth]{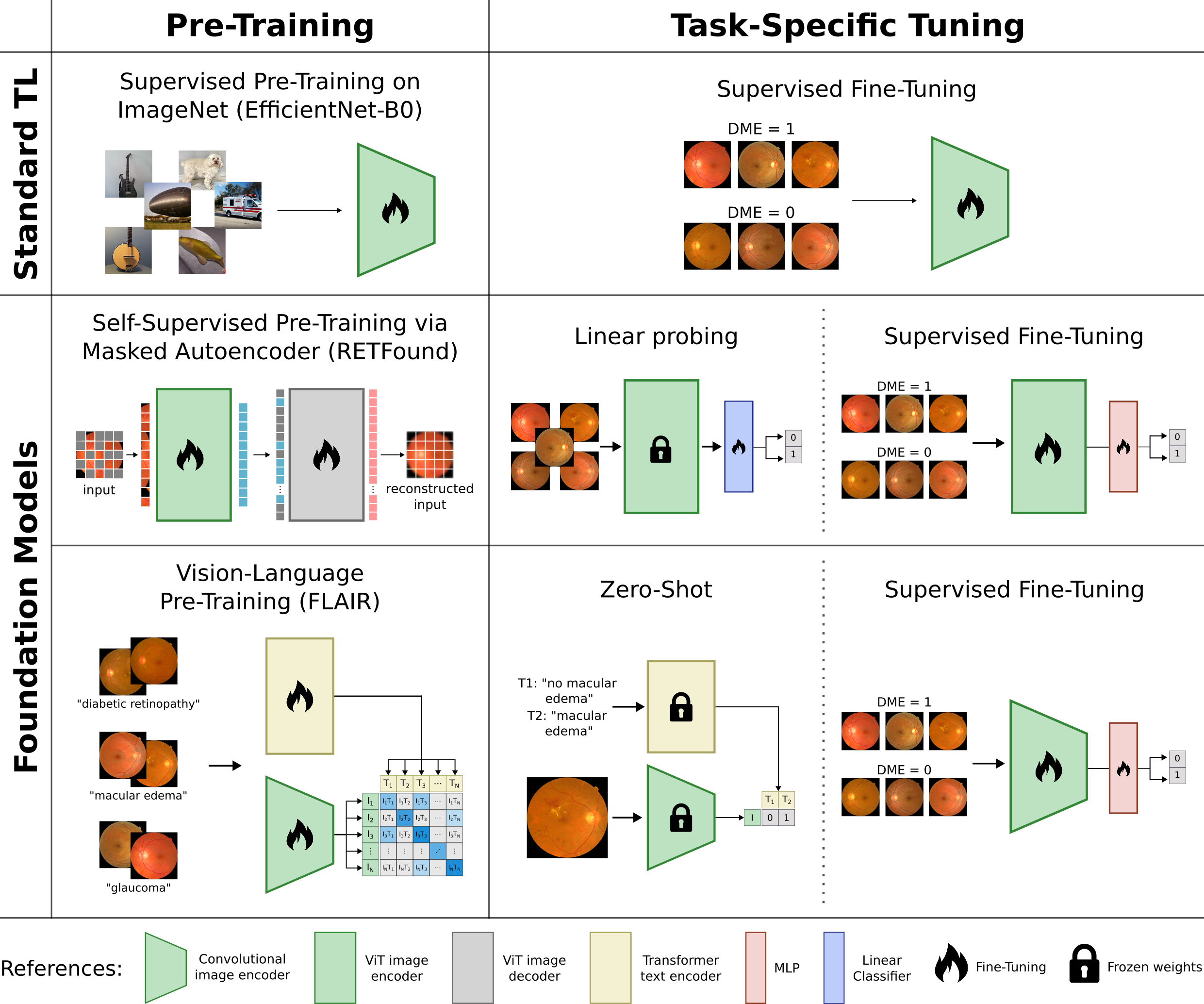}
  \caption{Schematic representation of our study. We compared the standard transfer learning (TL) approach of supervised task-specific fine-tuning of a CNN pre-trained on ImageNet against linear probing, supervised fine-tuning and zero-shot prediction using Foundation Models (FMs).}
  
  \label{fig:teaser}
\end{figure}

Diabetic macular edema (DME) is one of the most serious vision-threatening conditions linked to diabetic retinopathy (DR)~\cite{DME-definition}. It occurs when fluid accumulates in the macula, 
leading to swelling that distorts central sight~\cite{DME-fluid}. Early detection is crucial to initiate treatment promptly and prevent irreversible vision loss~\cite{DME-early-det}. But, as the disease progresses without noticeable symptoms, it may go undetected until patient experiences vision impairment~\cite{musat2015diabetic}, so regular checkups are recommended i.e. through fundus imaging, a time-consuming process that depends on ophthalmologists skills~\cite{xu2019diagnostic}.

In recent years, deep learning has shown strong potential for automating medical image analysis~\cite{deep-learning-medicine}, including the detection of retinal diseases~\cite{deep-learning-retina}. However, developing high-performing models for DME detection remains particularly challenging due to the limited availability of annotated data~\cite{limited-data-dme}. Foundation models (FMs) have emerged as a promising approach to address this limitation~\cite{foundation-models-medicine}. These models are pre-trained on large general-purpose datasets using pretext tasks--either via self-supervised learning from unlabeled samples~\cite{balestriero2023cookbook, retfound} or by learning to match images and their associated clinical reports. Such strategies help the models learn meaningful representations of the data, serving as a robust base for developing specialized models for downstream tasks through linear probing or fine-tuning on smaller, task-specific datasets~\cite{fm-downstream}.

To ensure alignment between the data used during pre-training and that used for fine-tuning, several domain-specific FM trained solely on retinal images have been introduced, such as RETFound~\cite{retfound}, FLAIR~\cite{flair}, and others~\cite{retclip, yu2024urfound, yang2024vilref}. RETFound, for example, employs self-supervised learning with masked autoencoders~\cite{masked-autoencoder} to learn an encoder that can be adapted to other fundus image-specific tasks. FLAIR~\cite{flair}, on the other hand, is trained to align image–text pairs using the CLIP framework~\cite{clip}. It combines an image encoder based on ResNet-50~\cite{resnet} with a text encoder based on BioClinicalBERT~\cite{bioclinicalbert}, enabling the development of a robust image encoder for fine-tuning, as well as a zero-shot model~\cite{zero-shot} capable of predicting outcomes by jointly processing image and text prompts and computing their cosine similarity~\cite{flair}.

Recent studies have explored the use of these models for detecting conditions such as hypertensive retinopathy~\cite{silva2023exploring}, DR~\cite{retclip, FM-test2}, glaucoma~\cite{retclip, FM-test2}, and others, demonstrating promising results. These include either superior performance compared to standard baselines~\cite{FM-test2, retfound, retclip} or improved data efficiency for downstream tasks~\cite{retfound}. However, the effectiveness of these models as a starting point for DME detection remains unexplored. In this study, we address this gap by evaluating the performance of FMs for DME detection under limited-data conditions~\cite{foundation-models-medicine} (Fig.~\ref{fig:teaser}). Specifically, we evaluate two of the most popular FMs--RETFound and FLAIR--under linear probing, fine-tuning, and/or zero-shot prediction scenarios, and compare them against an EfficientNet-B0~\cite{efficientnet} baseline pre-trained on ImageNet~\cite{deng2009imagenet} and fine-tuned for this task. Our results show that standard transfer learning remains a strong baseline for DME detection, often outperforming FM-derived models across multiple evaluation settings.

  \begin{table*}[t!]
  \centering
  \caption{AUC-PR/ROC results for different models trained and tested for DME detection. \textbf{First}, \underline{second} and \textit{third} ranked values are formatted accordingly. Values in brackets are 95\% confidence intervals obtained with bootstrapping ($n = 1000$).}
  \resizebox{0.99\textwidth}{!}{
  \begin{tabular}{l|lll|ccc|ccc}
  \toprule
  \multirow{2}{*}{\parbox{1.8cm}{\raggedright\textbf{Training set}}} &
  \multirow{2}{*}{\parbox{1.5cm}{\raggedright\textbf{Model}}} &
  \multirow{2}{*}{\parbox{3.2cm}{\raggedright\textbf{Architecture (parameters)}}} &
  \multirow{2}{*}{\parbox{2.2cm}{\raggedright\textbf{Setting}}} &
  \multicolumn{3}{c|}{\textbf{AUC-PR (95\% CI)}} & 
  \multicolumn{3}{c}{\textbf{AUC-ROC (95\% CI)}} \\
  \cmidrule(lr){5-7} \cmidrule(lr){8-10}
   & & & & \textbf{IDRiD} & \textbf{MESSIDOR-2} & \textbf{OEFI} & \textbf{IDRiD} & \textbf{MESSIDOR-2} & \textbf{OEFI} \\
  \midrule
  \multirow{7}{*}{\makecell[l]{IDRiD\\($N=371$)}}
    & CNN       & EfficientNet-B0 (5.3M) & SFT      & \textbf{0.966} {\tiny [0.93 - 0.99]} & \textbf{0.719} {\tiny [0.59 - 0.84]} & \underline{0.962} {\tiny [0.95 - 0.97]} & \textbf{0.953} {\tiny [0.89 - 0.99]} & \textbf{0.923} {\tiny [0.87 - 0.97]} & \textbf{0.941} {\tiny [0.93 - 0.95]} \\
    & RETFound  & ViT-S (307M)           & SFT      & \textit{0.920} {\tiny [0.86 - 0.96]} & \textit{0.497} {\tiny [0.34 - 0.63]} & \textit{0.956} {\tiny [0.95 - 0.96]} & \textit{0.897} {\tiny [0.82 - 0.96]} & \textit{0.814} {\tiny [0.74 - 0.88]} & \textit{0.913} {\tiny [0.90 - 0.93]} \\
    & RETFound  & ViT-S (307M)           & LP (Ridge)      & 0.607 {\tiny [0.47 - 0.74]} & 0.201 {\tiny [0.12 - 0.31]} & 0.697 {\tiny [0.66 - 0.73]} & 0.594 {\tiny [0.48 - 0.71]} & 0.666 {\tiny [0.57 - 0.75]} & 0.537 {\tiny [0.51 - 0.57]} \\
    & RETFound  & ViT-S (307M)          & LP (PCA + Ridge)& 0.608 {\tiny [0.47 - 0.74]} & 0.169 {\tiny [0.10 - 0.27]} & 0.688 {\tiny [0.66 - 0.72]} & 0.594 {\tiny [0.48 - 0.71]} & 0.641 {\tiny [0.55 - 0.72]} & 0.507 {\tiny [0.48 - 0.54]} \\
    & RETFound  & ViT-S (307M)          & LP (LASSO)      & 0.612 {\tiny [0.48 - 0.74]} & 0.186 {\tiny [0.11 - 0.29]} & 0.699 {\tiny [0.67 - 0.73]} & 0.606 {\tiny [0.49 - 0.72]} & 0.656 {\tiny [0.56 - 0.74]} & 0.516 {\tiny [0.48 - 0.55]} \\
    & RETFound  & ViT-S (307M)          & LP (PCA + LASSO)& 0.533 {\tiny [0.41 - 0.69]} & 0.250 {\tiny [0.16 - 0.41]} & 0.636 {\tiny [0.60 - 0.67]} & 0.585 {\tiny [0.47 - 0.69]} & 0.669 {\tiny [0.57 - 0.77]} & 0.350 {\tiny [0.32 - 0.38]} \\
    & FLAIR     & RN-50 (26M) & SFT      & \underline{0.930} {\tiny [0.87 - 0.97]} & \underline{0.686} {\tiny [0.55 - 0.80]} & \textbf{0.965} {\tiny [0.96 - 0.97]} & \underline{0.921} {\tiny [0.87 - 0.97]} & \underline{0.908} {\tiny [0.84 - 0.96]} & \underline{0.937} {\tiny [0.92 - 0.95]} \\
  \midrule
  \multirow{7}{*}{\makecell[l]{MESSIDOR-2\\($N=1096$)}}
    & CNN       & EfficientNet-B0 (5.3M) & SFT      & \textbf{0.956} {\tiny [0.91 - 0.99]} & \textbf{0.792} {\tiny [0.68 - 0.88]} & \underline{0.933} {\tiny [0.91 - 0.95]} & \textbf{0.940} {\tiny [0.87 - 0.98]} & \textbf{0.959} {\tiny [0.93 - 0.98]} & \textbf{0.918} {\tiny [0.90 - 0.94]} \\
    & RETFound  & ViT-S (307M) & SFT      & \textit{0.870} {\tiny [0.79 - 0.93]} & \textit{0.623} {\tiny [0.48 - 0.74]} & \textit{0.917} {\tiny [0.90 - 0.93]} & \textit{0.836} {\tiny [0.75 - 0.91]} & \textit{0.886} {\tiny [0.83 - 0.94]} & \textit{0.856} {\tiny [0.84 - 0.87]} \\
    & RETFound  & ViT-S (307M) & LP (Ridge)  & 0.514 {\tiny [0.39 - 0.68]} & 0.250 {\tiny [0.16 - 0.38]} & 0.554 {\tiny [0.53 - 0.58]} & 0.542 {\tiny [0.43 - 0.65]} & 0.784 {\tiny [0.71 - 0.85]} & 0.315 {\tiny [0.28 - 0.35]} \\
    & RETFound  & ViT-S (307M) & LP (PCA + Ridge)& 0.628 {\tiny [0.49 - 0.77]} & 0.203 {\tiny [0.13 - 0.33]} & 0.845 {\tiny [0.82 - 0.87]} & 0.631 {\tiny [0.51 - 0.74]} & 0.736 {\tiny [0.65 - 0.81]} & 0.756 {\tiny [0.73 - 0.78]} \\
    & RETFound  & ViT-S (307M) & LP (LASSO) & 0.576 {\tiny [0.44 - 0.73]} & 0.260 {\tiny [0.16 - 0.39]} & 0.574 {\tiny [0.55 - 0.60]} & 0.628 {\tiny [0.51 - 0.73]} & 0.786 {\tiny [0.72 - 0.85]} & 0.370 {\tiny [0.34 - 0.40]} \\
    & RETFound  & ViT-S (307M) & LP (PCA + LASSO)& 0.631 {\tiny [0.49 - 0.77]} & 0.204 {\tiny [0.13 - 0.33]} & 0.843 {\tiny [0.82 - 0.87]} & 0.634 {\tiny [0.51 - 0.75]} & 0.736 {\tiny [0.65 - 0.81]} & 0.755 {\tiny [0.73 - 0.78]} \\
    & FLAIR     & RN-50 (26M) & SFT      & \underline{0.909} {\tiny [0.83 - 0.95]} & \underline{0.703} {\tiny [0.56 - 0.81]} & \textbf{0.941} {\tiny [0.93 - 0.95]} & \underline{0.889} {\tiny [0.81 - 0.95]} & \underline{0.908} {\tiny [0.85 - 0.95]} & \underline{0.897} {\tiny [0.88 - 0.91]} \\
  \midrule
  \multirow{4}{*}{\makecell[l]{Zero-shot\\($N=0$)}}
    & FLAIR     & RN-50 (26M) / BCB (110M) & Prompt 1   & \textbf{0.849} {\tiny [0.75 - 0.92]} & \textbf{0.565} {\tiny [0.42 - 0.70]} & \underline{0.918} {\tiny [0.89 - 0.94]} & \underline{0.863} {\tiny [0.79 - 0.93]} & \textbf{0.936} {\tiny [0.91 - 0.96]} & \underline{0.907} {\tiny [0.89 - 0.93]} \\
    & FLAIR     & RN-50 (26M) / BCB (110M) & Prompt 2   & \textit{0.813} {\tiny [0.69 - 0.92]} & \textit{0.525} {\tiny [0.38 - 0.68]} & \textbf{0.945} {\tiny [0.93 - 0.96]} & \textit{0.856} {\tiny [0.78 - 0.93]} & 0.929 {\tiny [0.90 - 0.95]} & \textbf{0.935} {\tiny [0.92 - 0.95]} \\
    & FLAIR     & RN-50 (26M) / BCB (110M) & Prompt 3   & \underline{0.836} {\tiny [0.73 - 0.93]} & \underline{0.546} {\tiny [0.40 - 0.69]} & 0.861 {\tiny [0.83 - 0.89]} & \textbf{0.865} {\tiny [0.79 - 0.93]} & \underline{0.932} {\tiny [0.90 - 0.95]} & 0.792 {\tiny [0.77 - 0.82]} \\
    & FLAIR     & RN-50 (26M) / BCB (110M) & Prompt 4   & 0.794 {\tiny [0.68 - 0.91]} & 0.503 {\tiny [0.37 - 0.68]} & \textit{0.896} {\tiny [0.87 - 0.92]} & 0.848 {\tiny [0.77 - 0.92]} & \textit{0.931} {\tiny [0.90 - 0.95]} & \textit{0.868} {\tiny [0.85 - 0.89]} \\
  \bottomrule
  \end{tabular}}
  \label{tab:aucpr_roc_results}
  \captionsetup{justification=justified,singlelinecheck=false}
  \caption*{\footnotesize%
  \textit{RN-50}: ResNet-50.\quad
  \textit{BCB}: BioClinicalBERT.\quad
  \textit{SFT}: Supervised Fine-Tuning.\quad
  \textit{LP}: Linear Probing.\quad
  \textit{Prompt 1}: "macular edema".\quad
  \textit{Prompt 2}: "leakage of fluid within the central macula from microaneurysms".\quad
  \textit{Prompt 3}: "presence of exudates".\quad
  \textit{Prompt 4}: "presence of exudates within the radius of one disc diameter from the macula center".%
  }
  \end{table*}

\section{Methods}

Our study is schematically presented in Fig.~\ref{fig:teaser}. Specifically, we evaluate three methods: standard fine-tuning (Section~\ref{sec:methods-stl}), linear probing using FMs as fixed feature extractors (Section~\ref{sec:methods-lp}), and zero-shot prediction with vision-language FMs (Section~\ref{sec:methods-zs}). Each approach is described in the sequel.

\subsection{Standard Fine-Tuning}
\label{sec:methods-stl}
In the standard fine-tuning approach, we evaluate three models: EfficientNet-B0, the ResNet-50 encoder from FLAIR, and RETFound, which uses a Vision Transformer (ViT-S)~\cite{dosovitskiy2020image}. We chose EfficientNet-B0 pre-trained on ImageNet, as it is an established baseline for transfer learning. We also included the ResNet-50 from FLAIR to study the benefits of prior training on retinal images and text, which should enable more domain-specific representations. Finally, we included RETFound's ViT to assess the effect of using self-supervised learning.

Each network is adapted for DME detection by supervised fine-tuning (SFT) on retinal fundus images labeled as DME or non-DME. At this phase, all model layers are updated using the training data from each dataset, allowing them to specialize their representations for the DME detection task.

\subsection{Linear Probing}
\label{sec:methods-lp}

To explore the representational power of FMs, we evaluate linear probing as an alternative to full fine-tuning. In this setup, the weights of the backbone model remain frozen, and only a lightweight linear classifier is trained on top of the extracted features. We use RETFound as fixed feature extractor and train two classifiers--standard Ridge Regression and LASSO~\cite{tibshirani1996regression}--on its embeddings, to distinguish between DME and non-DME cases. To mitigate the effect of the so-called curse of dimensionality, we also trained these classifiers using dimensionality reduction via Principal Component Analysis (PCA).

\subsection{Zero-shot prediction}
\label{sec:methods-zs}

We also used FLAIR as a zero-shot classifier, leveraging its  abilities to align visual and textual representations~\cite{flair,clip}. In this setup, no fine-tuning are performed on networks' parameters. Instead, the model receives a fundus image along with predefined text prompts describing both presence/absence of DME-related findings. Each image is encoded by the ResNet-50 image encoder, and each prompt is processed by the BioClinicalBERT text encoder. Classification is then performed by computing the cosine similarity between image and text embeddings, and the predicted class is assigned based on the prompt with the highest similarity. No systematic prompt engineering was applied for evaluation. Instead, we evaluated the positive prompts describing DME-related findinds that were originally used for training FLAIR~\cite{flair} (Table~\ref{tab:aucpr_roc_results}), and constructed corresponding negative prompts by negating them. 

\section{Experimental setup}

\subsection{Materials}


Empirical evaluation was performed using three public datasets: the popular MESSIDOR-2~\cite{messidor,abramoff2013automated} and IDRiD~\cite{idrid} sets, widely applied for this task \cite{sundaram2023diabetic, wang2022diabetic, nazir2021detection}, and the recently introduced OCT-and-Eye-Fundus-Images (OEFI) set~\cite{oefi}. 

MESSIDOR-2 contains 1740 images labeled for DME as either negative (0) or positive (1) (1589 vs. 151, respectively). As no fixed training, validation and test partitions are provided, we randomly divided them using 70\% and 30\% for training and test, respectively, pulling off 10\% of the training set for validation. Stratified sampling at a patient level was used to ensure similar distribution between classes.

IDRiD, on the other hand, consists of 516 images, annotated with three DME grades (0--no DME--, 1--non-clinically significant DME--, or 2--DME--; 222, 51 and 243 images, respectively). \textcolor{black}{Labels 0 and 1 were merged into a single negative class, while label 2 was considered the positive class, resulting in a binary classification task.}  We followed the same partitions into training and test as provided in the set, while extracted 10\% of the training samples for validation.

Finally, OEFI includes 1548 eye fundus and 1113 OCT images acquired from multiple ophthalmological institutions in Mexico, all with binary DME labels. We only used the fundus images (1053 and 495 with and without DME, respectively), as an external set to study models' generalization performance.

\subsection{Evaluation metrics}
All models were evaluated using the Area Under the ROC (AUC-ROC) and Precision-Recall (AUC-PR) curves, two standard metrics in the literature for DME detection~\cite{bressler2022autonomous,retclip,retfound}. Given the noticeable class imbalance in some of the evaluation datasets--such as in MESSIDOR-2 (8.7\% positive cases) and OEFI (32.0\%),while IDRiD is more balanced 
\textcolor{black}{ (47.0\% positive)--}, we used AUC-PR on the validation set to choose the best configuration of each model. 

\subsection{Training configuration}

Supervised fine-tuning was performed using Adam optimization, with learning rates empirically chosen per model and dataset. A custom version of RandAugment~\cite{randaugment} was used for data augmentation. A grid search procedure was applied to fix the number of transformations (from 1 to 7) and augmentation strength (from 0.2 to 1.0, with increments of 0.2), choosing operations from a pool that included adjustments to brightness, contrast, saturation, and hue, random rotation, scaling, horizontal flipping, and Gaussian blur. Parameter ranges were empirically chosen to preserve clinical plausibility.

For linear probing, we trained both Ridge and LASSO regression models on fixed feature vectors obtained using the RETFound backbone. Each method was evaluated both with and without dimensionality reduction using PCA, retaining 99\% of the variance in the feature space when applied. The regularization parameter $\alpha$ was selected via grid search. For Ridge regression, we explored a broad range of $\alpha$ values: [0.001, 0.01, 0.1, 0.5, 1.0, 5.0, 10.0, 50.0, 100.0, 200.0], which allowed us to assess the effect of both weak and strong regularization. In contrast, for LASSO regression, we used a narrower and lower range: [0.0001, 0.001, 0.01, 0.1, 0.5, 1.0], as LASSO tends to drive coefficients to zero more aggressively and can underfit with higher regularization strengths~\cite{tibshirani1996regression}.

\section{Results}

All quantitative results are summarized in Table~\ref{tab:aucpr_roc_results}. For models trained on IDRiD, EfficientNet-B0 achieves the highest AUC-PR and AUC-ROC on the IDRiD test set. 
\textcolor{black}{Among FMs, fine-tuning FLAIR is the most competitive approach, followed by fine-tuning RETFound. On OEFI, FLAIR reports the highest AUC-PR, followed by EfficientNet-B0, whereas EfficientNet-B0 achieves the highest AUC-ROC, with FLAIR ranking second. In OEFI, the fine-tuned RETFound ranks third. Surprisingly, when these models are evaluated on MESSIDOR-2, a noticeable drop in performance is observed. This drop is particularly evident in the AUC-PR scores, where EfficientNet-B0 retains the top position but with a reduced maximum value of 0.719, while both FMs fall below 0.7.}

When models are trained on MESSIDOR-2, EfficientNet-B0 again achieves the highest AUC-PR and AUC-ROC on both MESSIDOR-2 and IDRiD test sets. The second-best performing model is the fine-tuned FLAIR, although with a larger performance gap than observed when training on IDRiD. The fine-tuned RETFound consistently ranks third by a significant margin. A similar pattern is observed when evaluated on OEFI, although in this case the fine-tuned FLAIR reports slightly higher AUC-PR values than EfficientNet-B0.

All linear probing strategies using RETFound embeddings--including Ridge, LASSO, and their PCA variants--yield very low AUC-PR and AUC-ROC values compared to fully supervised fine-tuning, regardless of the training set.
\textcolor{black}{The impact of PCA for dimensionality reduction is highly inconsistent and dataset-dependent. For models trained on IDRiD, PCA generally degrades performance, except for a notable improvement on MESSIDOR-2 using LASSO. Conversely, training on MESSIDOR-2 with PCA substantially improves cross-dataset evaluations (IDRiD and OEFI) for both regression methods, but consistently degrades in-domain results.}




In the zero-shot setting with FLAIR, the best AUC-PR on the IDRiD test set is achieved using Prompt 1. However, when evaluated using AUC-ROC, Prompt 3 yields the highest score. Conversely, on MESSIDOR-2, Prompt 1 convincingly performed best across both metrics. On the OEFI test set, the best performance is also consistently achieved with Prompt 2, which reports the highest AUC-PR and AUC-ROC values.

\textcolor{black}{When comparing zero-shot performance of FLAIR against the fine-tuned version of its ResNet-50 image encoder, fine-tuning significantly improves AUC-PR on both the IDRiD and MESSIDOR-2 test sets. While the zero-shot model reports slightly higher AUC-ROC scores on MESSIDOR-2, fine-tuning remains strictly superior when considering the more critical AUC-PR metric. On the OEFI test set, however, the advantage of fine-tuning diminishes; the best zero-shot prompt closely matches the model trained on IDRiD and even outperforms the version fine-tuned on MESSIDOR-2.}

\begin{figure}[t!]
  \centering
  \includegraphics[width=0.99\columnwidth]{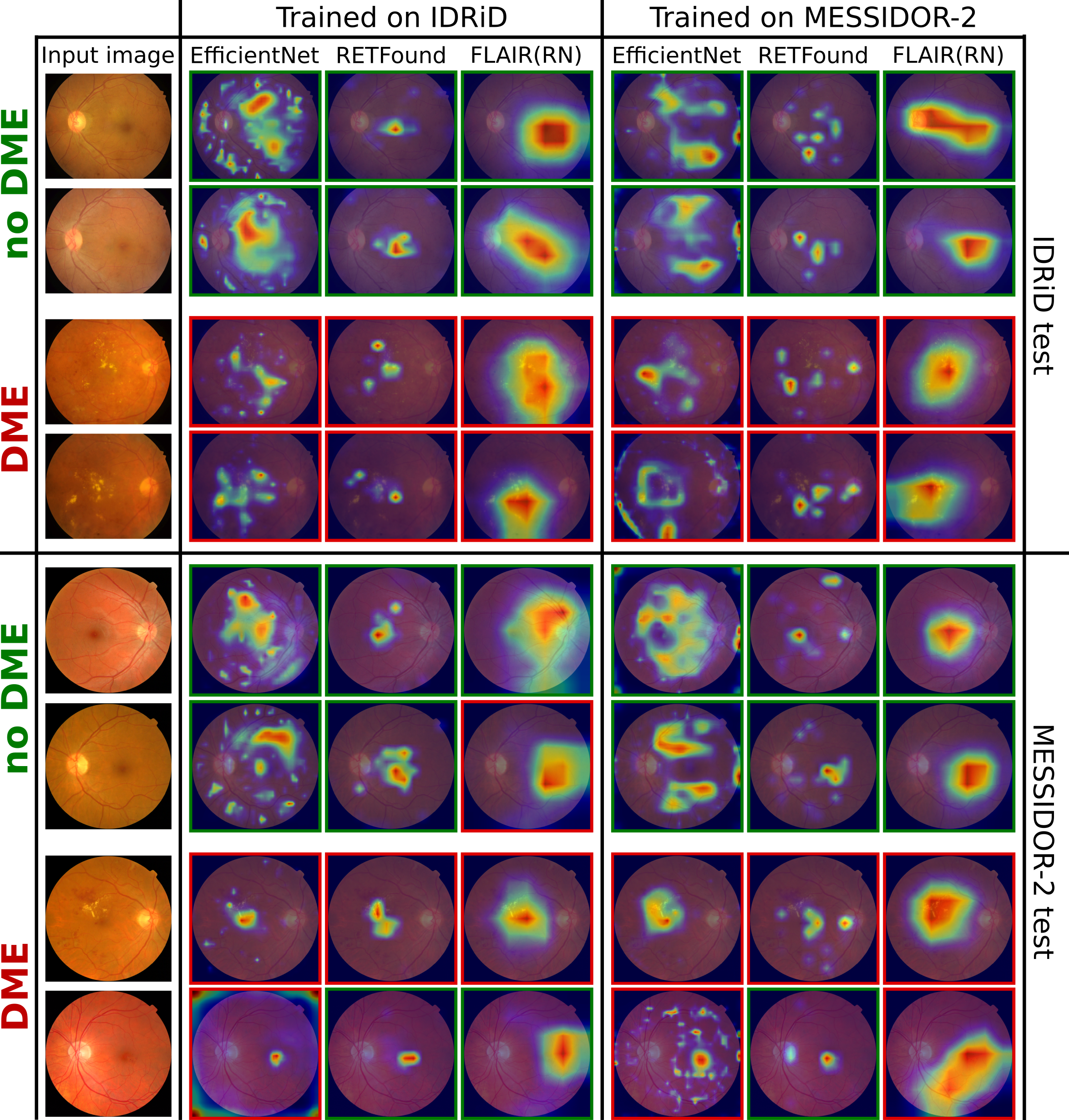}
  \caption{Qualitative comparison of explainability maps across models and datasets, for randomly selected IDRiD and MESSIDOR test images (top and bottom blocks), with and without DME. Heatmaps for EfficientNet-B0 and FLAIR (ResNet-50) are Grad-CAMs of the predicted class, while RETFound maps are Gradient Attention Rollouts. Border color indicates the model’s predicted label (green = no DME, red = DME).}
  \label{fig:heatmaps}
\end{figure}

Qualitative results are provided in Fig.~\ref{fig:heatmaps}, using explainability maps such as GradCAMs (for the CNNs) and Gradient Attention Rollout~\cite{abnar2020quantifying} (for RETFound).  
\textcolor{black}{
For positive cases, EfficientNet-B0 models successfully highlighted hard exudates, with relatively localized activations around pathological regions. When evaluating a subtle DME image (bottom row), both models classified it well but yielded noisy activations: the IDRiD-trained network showed activations along the image borders, while the MESSIDOR-2-one produced scattered activations across the retina. 
RETFound, on the other hand, displayed less consistent saliency: although it frequently activated around exudates in DME images. FLAIR, finally, generated diffuse activation maps consistently centered on the macula regardless of the training set, suggesting that it relies on lesion presence or absence in that region to guide its predictions.}

\section{Discussion}
\label{sec:discussion}

In this study we benchmarked two FMs for retinal image analysis--RETFound and FLAIR--for DME detection, comparing them with the typical approach of fine-tuning a CNN pre-trained on ImageNet. We selected these two due to their increased popularity and usage in the field~\cite{zhang2024retfound, sun2025medical}.

Existing literature claims that high-capacity FMs outperform lighter models on downstream tasks because pre-training endows them with rich, transferable representations requiring little additional refinement. Our results challenge this assumption. A lightweight CNN such as EfficientNet-B0 (with $\approx$5 M parameters) pre-trained on ImageNet and fine-tuned on fundus images consistently matched--or even surpassed-- the performance of much larger FM backbones such as the ViT-S encoder of RETFound ($\approx$307 M parameters) and the ResNet-50 image encoder of FLAIR ($\approx$26 M parameters). 
\textcolor{black}{We hypothesize that DME detection depends on fine-grained, local cues, such as small exudates and their spatial relationship to the macula, for which the convolutional inductive biases and task-specific fine-tuning of EfficientNet-B0 may be particularly well suited. FLAIR, which is also convolutional, remains the most competitive FM across the evaluated settings, consistently outperforming RETFound. This advantage may stem also from its language-supervised pre-training. Qualitative heatmaps in Fig.~\ref{fig:heatmaps} support this notion: EfficientNet-B0 produces more localized activations around clinically relevant pathological regions, whereas FLAIR tends to focus more broadly on the retinal region and RETFound appears to miss some small exudates.}

Our experiments also show that linear probing on RETFound embeddings yields markedly lower performance than full fine-tuning, regardless of dataset or classifier, indicating that successful transfer to specialized tasks such as DME detection requires deeper adaptation--updating internal representations, not merely the classifier head. 
This gap likely arises because RETFound’s self-supervised pretraining captures broad retinal structures rather than the subtle, localized biomarkers of DME, making the representations insufficiently discriminative for linear probing without deeper adaptation.


\textcolor{black}{
FLAIR demonstrates modest zero-shot capabilities, which is expected given that its vision-language pre-training already aligned retinal images with relevant clinical concepts. However, supervised fine-tuning is still necessary to achieve optimal performance in terms of AUC-PR. While the zero-shot approach occasionally yields higher AUC-ROC scores, this language-guided potential is bottlenecked by a strong dependence on prompt wording, as no single prompt guarantees stable performance across all datasets. Further work is needed to clarify how prompt phrasing and dataset characteristics interact to affect zero-shot accuracy.}

Beyond these model-specific findings, we consistently observed lower performance on MESSIDOR-2 compared to IDRiD. This drop is likely driven by domain shift between the datasets: IDRiD images are high-resolution, captured with dilated pupils and systematically annotated for exudates near the macula, whereas MESSIDOR-2 comprises routine clinical acquisitions with lower resolution, variable illumination, and a much lower prevalence of DME cases. These discrepancies in acquisition protocols, image quality, and disease distribution make cross-dataset generalization particularly challenging, explaining the reduced transferability observed across all models.

Our study shows that FMs are not a one-size-fits-all solution for DME detection. 
\textcolor{black}{Despite their capacity, the evaluated FMs did not outperform a lightweight EfficientNet-B0. FLAIR consistently outperforms RETFound among evaluations, suggesting that differences in pre-training  and architecture may influence their suitability for fine-grained ophthalmic tasks. These results highlight the importance of task-specific adaptation and careful benchmarking against conventional baselines, specially under limited-data regimes. We encourage future work to compare these findings against new FMs or even large, generalistic~\cite{tomita2025image,liang2025novel} or specialized VLMs~\cite{sellergren2025medgemma}.}

\section*{Acknowledgments}
This study was partially funded with a CONICET PIP 2021-2023 (11220200102472CO) and FinExa funds (Facultad de Ciencias Exactas - UNICEN).

\ifCLASSOPTIONcaptionsoff
  \newpage
\fi

\bibliographystyle{IEEEtran}
\bibliography{references.bib}

\end{document}